\pdfoutput=1

\documentclass[11pt]{article}

\usepackage[]{acl}

\usepackage{times}
\usepackage{latexsym}

\usepackage[T1]{fontenc}

\usepackage[utf8]{inputenc}

\usepackage{microtype}

\usepackage{booktabs}
\usepackage{multirow}
\usepackage{comment}
\usepackage{subcaption}
\usepackage{graphicx}
\usepackage{cleveref}
\usepackage{url}

\newcommand{\cpara}{$x_\mathrm{T}^\prime$}
\newcommand{\ppara}{$x_\mathrm{T}^{\prime\prime}$}

\newcommand{\lonesim}{$\mathcal{L}_\mathrm{1}$}

\newcommand{\ltwosim}{$\mathcal{L}_\mathrm{2}$}
\newcommand{\lthreesim}{$\mathcal{L}_\mathrm{3}$}

\title{Fine-tuning CLIP Text Encoders with Two-step Paraphrasing}

\author{Hyunjae Kim$^1$ \quad Seunghyun Yoon$^2$ \quad Trung Bui$^2$  \\ \textbf{\quad Handong Zhao$^2$} \quad 
\textbf{Quan Tran$^2$} \quad \textbf{Franck Dernoncourt$^2$} \quad \textbf{Jaewoo Kang$^{1}$}\\
$^1$Korea University \quad $^2$Adobe Research \\
\texttt{\{hyunjae-kim,kangj\}@korea.ac.kr} \\ 
\texttt{\{syoon,bui,hazhao,qtran,dernonco\}@adobe.com}
}

\begin{document}
\maketitle

\begin{abstract}
Contrastive language-image pre-training (CLIP) models have demonstrated considerable success across various vision-language tasks, such as text-to-image retrieval, where the model is required to effectively process natural language input to produce an accurate visual output. 
However, current models still face limitations in dealing with linguistic variations in input queries, such as paraphrases, making it challenging to handle a broad range of user queries in real-world applications.
In this study, we introduce a straightforward fine-tuning approach to enhance the representations of CLIP models for paraphrases.
Our approach involves a two-step paraphrase generation process, where we automatically create two categories of paraphrases from web-scale image captions by leveraging large language models. 
Subsequently, we fine-tune the CLIP text encoder using these generated paraphrases while freezing the image encoder.
Our resulting model, which we call ParaCLIP, exhibits significant improvements over baseline CLIP models across various tasks, including paraphrased retrieval (with rank similarity scores improved by up to 2.0\% and 5.6\%), Visual Genome Relation and Attribution, as well as seven semantic textual similarity tasks.
\end{abstract}
\section{Introduction}


Contrastive language-image pre-training (CLIP) models~\cite{radford2021learning} have gained significant attention in the fields of computer vision and natural language processing for their remarkable capacity to understand the relationship between text and images. 
They have been widely used in various vision-language applications, including image classification~\cite{deng2009imagenet}, image retrieval~\cite{lin2014microsoft,plummer2015flickr30k}, and text-to-image generation~\cite{saharia2022photorealistic,rombach2022high}, where the model should return desired visual outputs for a given text, and vice versa.

\begin{figure}[t]
\centering
\begin{subfigure}{.99\columnwidth}
  \centering
  \includegraphics[width=.99\linewidth]{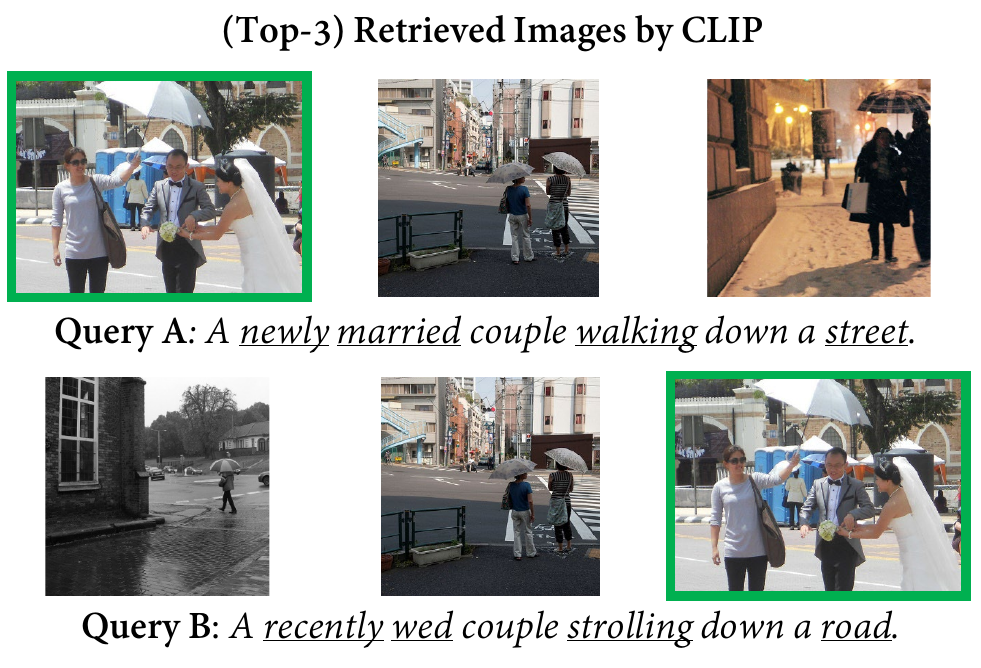} 
\end{subfigure}
\caption{
Image retrieval results of CLIP~\cite{radford2021learning} for two different queries (the gold image is denoted by a bold border). 
Despite their comparable meanings, the model yields dissimilar retrieval results, highlighting the model's struggle with linguistic variations.
}
\label{fig:motivating}
\end{figure}

An inherent challenge in vision-language tasks lies in the variability of text inputs. 
Even when conveying similar meanings and intentions, they can exhibit variations in vocabulary and structure depending on the particular user.
Consequently, it becomes crucial to ensure that CLIP's text encoders are robust enough to handle diverse synonyms and paraphrases in practical scenarios.
However, current text encoders exhibit limited proficiency in comprehending linguistic variations, resulting in different retrieval results for user queries with similar meanings (\Cref{fig:motivating}).

\begin{figure*}[t]
\centering
\begin{subfigure}{.99\linewidth}
  \centering
  \includegraphics[width=.99\linewidth]{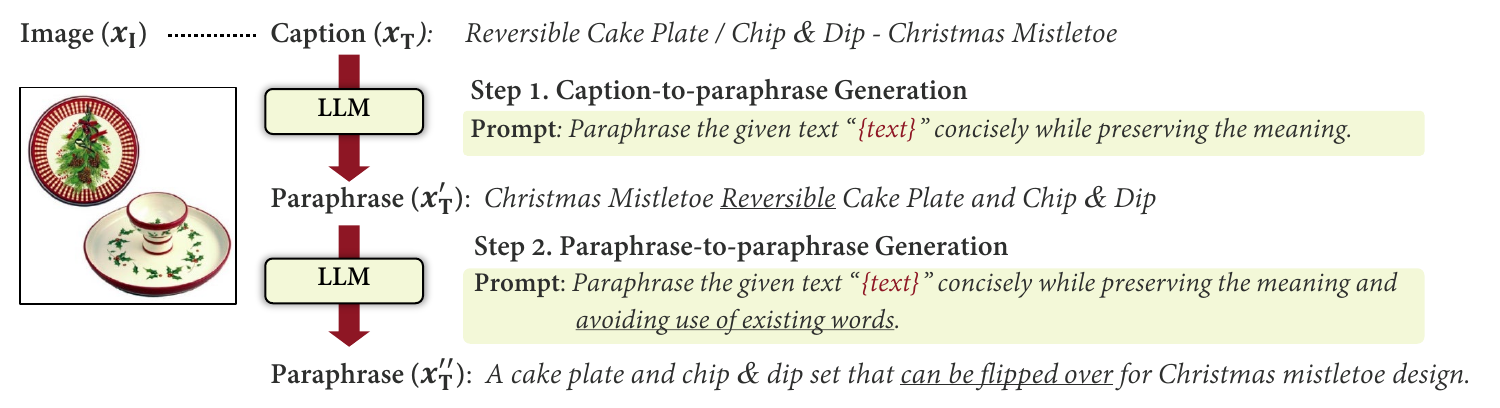} 
\end{subfigure}
\caption{
Overview of our two-step paraphrasing process. 
(1) In caption-to-paraphrase generation, the first paraphrase is generated by removing noise from the original caption and converting it into a more plain language.
(2) In paraphrase-to-paraphrase generation, the second paraphrase is generated from the first paraphrase, where the word ``reversible'' is changed to a semantically similar expression ``can be flipped over.''
}
\label{fig:data_generation}
\end{figure*}

To address this challenge, we introduce a straightforward method to improve CLIP's text encoders.
Specifically, we generated two categories of paraphrases for image captions sourced from the web, leveraging recent large language models (LLM) such as ChatGPT~\cite{openai2022chatgpt} and LLaMA~\cite{touvron2023llama}.
Subsequently, we utilized image captions and their corresponding paraphrases to fine-tune the text encoder, which ensures that the representations of captions and paraphrases cluster in a similar vector space.

We validated the effectiveness of our approach using evaluation tasks that assess models' understanding of language semantics and composition: paraphrased retrieval, Visual Genome Relation (VG-R), Visual Genome Attribution (VG-A)~\cite{yuksekgonul2022and}, and semantic textual similarity (STS) tasks~\cite{agirre-etal-2012-semeval}.
Our models, ParaCLIP, significantly outperformed baseline CLIP models, while maintaining or sometimes improving its robust performance on zero-shot image classification~\cite{deng2009imagenet}, as well as text and image retrieval~\cite{lin2014microsoft}.
We emphasize that this is the first study to improve the representations of CLIP's text encoders during the fine-tuning stage using synthetic paraphrases.

\section{Method}
Our objective is to refine the CLIP model's training process, enabling its text encoder to produce consistent representations for various semantically similar textual inputs that the model might encounter in real-world scenarios.
Certain image-captioning datasets provide multiple captions for a single image~\cite{lin2014microsoft,plummer2015flickr30k}, which might be utilized as semantically similar text pairs during training. 
However, the volume of these datasets is limited, which presents a challenge in terms of exposing models to diverse language patterns.
Therefore, we automatically generated semantically similar pairs (i.e., paraphrases) for millions of image captions sourced from the web.

\subsection{Paraphrase Generation}
\label{subsec:paraphrase_gen}

An image-captioning dataset typically comprises a collection of image-caption pairs $(x_\mathrm{I}, x_\mathrm{T})$, where $x_\mathrm{I}$ and $x_\mathrm{T}$ represent an image and the corresponding caption, respectively.
For each caption $x_\mathrm{T}$, we created two categories of paraphrases through a two-step paraphrasing process, caption-to-paraphrase generation and paraphrase-to-paraphrase generation, as illustrated in \Cref{fig:data_generation}.

\paragraph{Caption-to-paraphrase generation}

This process directly rewrites original captions.
Image captions on the web often contain considerable noise, such as superfluous punctuation, product codes, and file extensions, which differ from typical queries.
This step can be seen as responsible for converting these noisy captions into a more straightforward text format commonly used in everyday language.
Using the power of LLMs, we synthesized paraphrases \cpara~for each caption with the following prompt:
\textit{``Paraphrase the given caption ``\texttt{text}'' concisely while preserving the meaning.''},
where \texttt{text} is substituted with a given caption.

\paragraph{Paraphrase-to-paraphrase generation}

In this step, additional paraphrases, \ppara, are generated for each generated paraphrase, \cpara. 
The paraphrasing process is similar to the previous step, but with some differences in the prompt as follows:
\textit{``Paraphrase the given text ``\texttt{text}'' concisely while preserving the meaning and \underline{avoiding use of existing words.}''},
where the underlined text is used to prompt the model to produce morphologically diverse expressions.

\subsection{Training Objectives}
\label{subsec:objectives}

Let $\mathbf{X}_\mathrm{I}$, $\mathbf{X}_\mathrm{T}$, $\mathbf{X}_\mathrm{T}^\prime$, and $\mathbf{X}_\mathrm{T}^{\prime\prime}$ be
mini-batches of $N$ examples of an image $x_\mathrm{I}$, caption $x_\mathrm{T}$, and two types of paraphrases, \cpara~and \ppara.
The final loss is calculated as the summation of three sub-losses as follows:
$
\mathcal{L}_\mathrm{total} := \mathcal{L}_\mathrm{1}(\mathbf{X}_\mathrm{I},\mathbf{X}_\mathrm{T}^{\prime\prime}) + \mathcal{L}_\mathrm{2}(\mathbf{X}_\mathrm{T},\mathbf{X}_\mathrm{T}^{\prime}) + \mathcal{L}_\mathrm{3}(\mathbf{X}_\mathrm{T}^{\prime},\mathbf{X}_\mathrm{T}^{\prime\prime}).
$
The first term, $\mathcal{L}_\mathrm{1}$, represents the InfoNCE loss function that operates between images and text~\cite{oord2018representation}.
This loss function is crucial in the prevention of forgetting CLIP's representations and knowledge acquired during pre-training.
We used the paraphrased version of text input $\mathbf{X}_\mathrm{T}^{\prime\prime}$ rather than the original captions $\mathbf{X}_\mathrm{T}$ because user queries often resemble plain text rather than the original captions.
This choice led to improved performance on the benchmark datasets during our preliminary experiment.
If the target domain involves dealing with noisy text inputs, such as in an online shopping mall context, employing the original captions may be more effective.

The second term, $\mathcal{L}_\mathrm{2}$, accounts for the relationship between captions and their paraphrases.
Conceptually, it serves to establish a connection within the vector space between the representation of noisy captions and the plain text commonly used in everyday language.
Lastly, $\mathcal{L}_\mathrm{3}$ serves to bring together various semantically similar plain texts within a vector space. 
For $\mathcal{L}_\mathrm{2}$ and $\mathcal{L}_\mathrm{3}$, we used the InfoNCE loss.
The resulting CLIP model fine-tuned using these three losses is called ParaCLIP.

\section{Experimental Setups}

We obtained image-caption pairs using LAION-400M~\cite{schuhmann2021laion}.
We initially generated 300K paraphrases using ChatGPT and instruction-tuned an open-sourced LLM named LLaMA (7B)~\cite{touvron2023llama} using these 300K data to generate additional paraphrases.\footnote{We verified that the data generated by LLaMA exhibited comparable quality to that of ChatGPT. Additionally, when training the model using 300K paraphrases from LLaMA and an additional 300K paraphrases from ChatGPT, respectively, we observed similar performance in both cases.}
Our final dataset comprises 5M examples of $x_\mathrm{I}$, $x_\mathrm{T}$, \cpara, and \ppara.
More details and hyperparameters are described in \Cref{appendix:details}.


\subsection{Baseline Models}

We used the following CLIP models as baseline models, all built upon the ViT-B/32 architecture~\cite{dosovitskiy2020image}.
(1) {OpenAI's CLIP}~\cite{radford2021learning} was trained using a private dataset comprising 400M image-text pairs sourced from the web.
(2) {OpenCLIP} models~\cite{cherti2023reproducible} were trained using the largest open-sourced datasets, LAION-400M and LAION-2B~\cite{schuhmann2022laion}.
(3) OpenCLIP-RoBERTa was pre-trained using LAION-2B.
In contrast to the usual practice where text encoders are initialized with random weights and subsequently trained from scratch, its text encoder was initialized with the weights of RoBERTa-base~\cite{liu2019roberta} for better linguistic comprehension capabilities.
(4) LaCLIP~\cite{fan2023improving} was pre-trained using the LAION-400m dataset augmented with automatically generated paraphrases.\footnote{https://github.com/LijieFan/LaCLIP}
Specifically, a small number of original caption and paraphrase pairs were obtained from COCO text descriptions, or created by ChatGPT, Google BARD, and humans.
These seed examples were used to prompt an LLaMA 7B model through a in-context learning approach, which then generated paraphrases for the entire LAION-400m dataset.
During pre-training, a standard InfoNCE loss was computed using these paraphrases and corresponding images in combination with original caption and image pairs.
While our method shares some similarities with LaCLIP in the use of model-generated paraphrases, it should be noted that ours has unique advantages.
First, we enhance CLIP models through fine-tuning the text encoders while freezing the image encoders, which is significantly more efficient compared to pre-training the entire model from scratch.
Despite its efficiency, our method is significantly more effective to improve the CLIP's robustness to paraphrases, improving the performance in paraphrased retrieval by a large margin (see~\Cref{sec:results} for details).

\begin{table*}[ht]
\centering
\footnotesize
\begin{tabular}{lccccc|ccc}
\toprule
\multirow{3}{*}{\begin{tabular}[c]{@{}l@{}} \textbf{Model} \end{tabular}} & \multicolumn{2}{c}{\textbf{Paraphrased Rtrv.}} & \multicolumn{1}{c}{\textbf{VG-R}} & \multicolumn{1}{c}{\textbf{VG-A}} & \textbf{STS} & \multicolumn{1}{c}{\textbf{Clsf.}} & \multicolumn{1}{c}{\textbf{T Rtrv.}} & \multicolumn{1}{c}{\textbf{I Rtrv.}}  \\
\cmidrule(lr){2-3} \cmidrule(lr){4-4} \cmidrule(lr){5-5} \cmidrule(lr){6-6} \cmidrule(lr){7-7} \cmidrule(lr){8-8} \cmidrule(lr){9-9}
 & \textbf{AO@10} & \textbf{JS@10} & \textbf{Acc} & \multicolumn{1}{c}{\textbf{Acc}} & \textbf{Avg.} & \textbf{Acc} & \multicolumn{1}{c}{\textbf{R@5}} & \textbf{R@5} \\
\midrule
OpenAI's CLIP (400M) & 67.2 & 57.7 & 59.7 & 63.2 & 65.1 & {63.4} & 75.0 & \multicolumn{1}{c}{54.8}  \\
\quad+ ParaCLIP & \textbf{72.2} & \textbf{63.3} & \textbf{60.7} & \textbf{64.3} & \textbf{72.2} & \textbf{63.5} & \textbf{77.0} & \textbf{58.8}   \\
\midrule
OpenCLIP (400M) & 67.6 & 58.9 & 46.4 & 57.8 & 67.2 & 60.2 & \textbf{76.5} & \textbf{59.4}  \\
\quad+ ParaCLIP & \textbf{71.3} & \textbf{62.9} & \textbf{55.4} & \textbf{61.7} & \textbf{70.1} & \textbf{60.8} & 76.1 & \textbf{59.4}  \\
\midrule
OpenCLIP (2B) & 70.6 & 62.1 & 45.0 & 61.8 & 69.6 & \textbf{66.5} & 80.2 & \textbf{64.8}  \\
\quad+ ParaCLIP & \textbf{73.2} & \textbf{65.1} & \textbf{58.8} & \textbf{65.4} & \textbf{71.6} & 65.5 & \textbf{80.4} & 63.3  \\
\midrule
OpenCLIP-RoBERTa (2B) & 72.5 & 64.0 & 35.6 & 64.5 & 71.0 &  \textbf{61.8} & 78.8 & \textbf{62.6}  \\
\quad+ ParaCLIP & \textbf{74.5} & \textbf{66.2} & \textbf{43.2} & \textbf{66.5} & \textbf{72.5} & 61.4 & \textbf{79.4} & 62.0  \\
\midrule
LaCLIP (400M) & 69.9 & 62.1 & 50.6 & 63.6 & 58.8 & \textbf{64.5} & 68.1 & 55.5  \\
\quad+ ParaCLIP & \textbf{73.5} & \textbf{65.8} & \textbf{60.6} & \textbf{64.6} & \textbf{71.4} & \textbf{64.5} & \textbf{73.6} & \textbf{58.0} \\
\bottomrule
\end{tabular}
\caption{Zero-shot performance of baseline CLIP models and our ParaCLIP models. 
The best scores are represented in bold.
``Acc'': Accuracy. ``Avg.'': Macro average of Spearman's rank correlations across all STS tasks. ``Clsf.'': Image classification. ``T Rtrv.'': Text retrieval. ``I Rtrv.'': Image retrieval.}
\label{tab:main_results}
\end{table*}

\subsection{Evaluation}
We evaluated models on the following tasks in a zero-shot manner, without fine-tuning them on the target tasks.
(1) {Paraphrased retrieval~\cite{cheng2024adapting}} involves retrieving identical images for both 4,155 original queries and their corresponding paraphrases from the image set of the COCO 2017 validation set~\cite{lin2014microsoft}.
Paraphrases were generated using GPT-3~\cite{brown2020language} and subsequently verified by humans.
This task is well-suited for assessing models' ability to effectively handle user queries expressed in diverse forms.
For metrics, we used the top-10 average overlap (AO@10) and Jaccard similarity (JS@10) scores, which measure the degree of rank similarity between the top 10 images retrieved for the original query and paraphrased query. 
Detailed descriptions of the metrics can be found in \Cref{appendix:metrics}.

(2) VG-R and (3) VG-A~\cite{yuksekgonul2022and} are devised to assess relational and attributive understanding of vision-language models, respectively.
They involve determining the correct caption for a given image from two candidate captions, where negative captions are generated by interchanging objects based on their relational context or interchanging attributes of objects.
For instance, given the correct caption ``the \textit{dog} is behind the \textit{tree},'' a negative counterpart could be formulated as follows: ``the \textit{tree} is behind the \textit{dog}.''
The VG-R and VG-A datasets comprise 23,937 and 28,748 test examples, respectively.

(4) STS has been widely employed to evaluate the text representations of encoders~\cite{conneau-etal-2017-supervised,reimers-gurevych-2019-sentence,chuang-etal-2022-diffcse}.
This task involves measuring semantic similarity or relatedness between pairs of text.
Following \citet{gao-etal-2021-simcse}, we measured Spearman’s correlation for each task in the ``all'' aggregation setting and reported macro-averaged scores across the seven STS tasks~\cite{agirre-etal-2012-semeval,agirre-etal-2013-sem,agirre-etal-2014-semeval,agirre-etal-2015-semeval,agirre-etal-2016-semeval,cer-etal-2017-semeval,marelli-etal-2014-sick}. 

Additionally, we assessed whether our models can maintain or even improve their performance on standard vision or vision-language tasks after being fine-tuned, including zero-shot image classification on the ImageNet-1K validation set~\cite{deng2009imagenet}, and image-to-text retrieval and text-to-image retrieval on the COCO validation set~\cite{lin2014microsoft}.
For metrics, top-1 accuracy (Acc) and top-5 recall (R@5) were used in the classification and retrieval tasks, respectively.

\section{Results and Discussion}
\label{sec:results}

\subsection{Main Results}

\Cref{tab:main_results} shows the zero-shot performance of the baseline and our models in the evaluation tasks.

\paragraph{Effect of fine-tuning using paraphrases}
Across all CLIP models, our approach consistently demonstrated improved performance in the four primary tasks. 
Notably, the most significant improvements were observed in the paraphrased retrieval task, where our ParaCLIP model achieved 72.2\% and 63.3\% in AO@10 and JS@10 scores, increasing the performance of OpenAI's CLIP by 5.0\% and 5.6\%, respectively.\footnote{A case study comparing CLIP and ParaCLIP in the paraphrased retrieval task can be found in \Cref{appendix:case_study}.}
The improvements in the STS tasks are also noticeable, with the macro-average score improving by 7.1\%.
Although not in all cases, our approach generally enhances performance in the text retrieval task. 
This is attributed to our model's capability to encode texts that shares semantic similarity with a given input image closely within the vector space.


\begin{table*}[ht]
\centering
\footnotesize
\begin{tabular}{lccccc|ccc}
\toprule
\multirow{3}{*}{\begin{tabular}[c]{@{}l@{}} \textbf{Model} \end{tabular}} & \multicolumn{2}{c}{\textbf{Paraphrased Rtrv.}} & \multicolumn{1}{c}{\textbf{VG-R}} & \multicolumn{1}{c}{\textbf{VG-A}} & \textbf{STS} & \multicolumn{1}{c}{\textbf{Clsf.}} & \multicolumn{1}{c}{\textbf{T Rtrv.}} & \multicolumn{1}{c}{\textbf{I Rtrv.}}  \\
\cmidrule(lr){2-3} \cmidrule(lr){4-4} \cmidrule(lr){5-5} \cmidrule(lr){6-6} \cmidrule(lr){7-7} \cmidrule(lr){8-8} \cmidrule(lr){9-9}
 & \textbf{AO@10} & \textbf{JS@10} & \textbf{Acc} & \multicolumn{1}{c}{\textbf{Acc}} & \textbf{Avg.} & \textbf{Acc} & \multicolumn{1}{c}{\textbf{R@5}} & \textbf{R@5} \\
\midrule
OpenAI's CLIP (400M) & 67.2 & 57.7 & 59.7 & 63.2 & 65.1 & {63.4} & 75.0 & \multicolumn{1}{c}{54.8}  \\
+ \lonesim & 68.9 & 59.9 & 58.0 & 62.4 & 68.7 & {63.7} & 75.8 & 58.0  \\
+ $\mathcal{L}_\mathrm{2} + \mathcal{L}_\mathrm{3}$ & 70.5 & 61.2 & \textbf{61.5} & \textbf{65.1} & \textbf{74.5} & 56.7 & 74.6 & 51.8  \\
+ $\mathcal{L}_\mathrm{1} + \mathcal{L}_\mathrm{1}^{\prime}$ & 70.4 & 61.7 & 58.2 & 63.0 & 69.1 & \underline{64.0} & 76.3 & \underline{58.7}  \\
+ $\mathcal{L}_\mathrm{1} + \mathcal{L}_\mathrm{1}^{\prime} + \mathcal{L}_\mathrm{1}^{\prime\prime}$ & \underline{71.3} & \underline{62.8} & 58.9 & 63.4 & 68.8 & \textbf{64.1} & \underline{76.4} & \textbf{58.8} \\
+ $\mathcal{L}_\mathrm{1} + \mathcal{L}_\mathrm{2}$ & 69.1 & 60.0 & 59.1 & 63.3 & 71.8 & 63.5 & 76.1 & 58.2 \\
+ $\mathcal{L}_\mathrm{1}^\prime + \mathcal{L}_\mathrm{2}$ & 70.8 & 62.0 & \underline{60.4} & 64.0 & 71.6 & {63.7} & \underline{76.4} & 58.6  \\
+ $\mathcal{L}_\mathrm{1} + \mathcal{L}_\mathrm{2} + \mathcal{L}_\mathrm{3}$ & 69.6 & 60.5 & 59.2 & 63.4 & \underline{72.4} & 63.1 & \underline{76.4} & 58.1 \\
+ $\mathcal{L}_\mathrm{1}^{\prime\prime} + \mathcal{L}_\mathrm{2} + \mathcal{L}_\mathrm{3}$ (\textbf{Ours}) & \textbf{72.2} & \textbf{63.3} & \underline{60.4} & \underline{64.2} & 72.2 & 63.5 & \textbf{77.0} & \textbf{58.8}  \\
\bottomrule
\end{tabular}
\caption{Zero-shot performance of OpenAI's CLIP (400M) with different loss functions applied. 
The best scores are represented in bold and the second best scores are underlined.
``Paraphrased Rtrv.'': Paraphrased retrieval. ``Acc'': Accuracy. ``Avg.'': Macro average of Spearman's rank correlations across all STS tasks. ``Clsf.'': Image classification. ``T Rtrv.'': Text retrieval. ``I Rtrv.'': Image retrieval.}
\label{tab:laclip_results}
\end{table*}

\paragraph{Effect of initialization with RoBERTa}
The OpenCLIP-RoBERTa model significantly outperformed the OpenCLIP (2B) model in paraphrased retrieval and STS, highlighting the benefits of leveraging pre-trained language models over randomly initialized text encoders. 
However, even with these advancements, there is substantial room for improvement in performance on these tasks.
Our fine-tuning approach further refined the RoBERTa text encoder, leading to notable achievements across the four primary tasks, with 2.0\% (AO@10) and 2.2\% (JS@10) scores in paraphrased retrieval.

\paragraph{Comparison with LaCLIP}

While LaCLIP exhibited superior performance compared to the OpenCLIP (400M) model in image classification, paraphrased retrieval, VG-R, and VG-A, its performance in the text/image retrieval and STS tasks witnessed a decline. 
This indicates that augmenting paraphrased text data may not consistently yield improvements, without incorporating effective loss functions such as \ltwosim~and \lthreesim.
Conversely, our fine-tuning method dramatically enhanced LaCLIP's performance in paraphrased retrieval (+ 3.6\% in AO@10 and 3.7\% in JS@10), VG-R (+ 10.0\%), VG-A (+ 1.0\%), STS (+ 12.6\%), and even on text retrieval (+ 5.5\%) and image retrieval (+ 2.5\%), highlighting that our method can complement LaCLIP to achieve optimal performance.

\paragraph{Lack of compositional understanding}
All CLIP models exhibited significant deficiencies in the VG-R and VG-A tasks. 
These limitations in compositional understanding can lead to errors in downstream tasks such as text-to-image synthesis, including unintentional attribute interchanges or the omission of objects in generated images~\cite{feng2023training}.
In future research, we plan to conduct a more in-depth analysis to explore the potential of our approach to mitigate these issues.

\subsection{Ablation Study}
\label{appendix:loss}

We conducted an ablation study to closely examine the individual contributions of each loss term (\Cref{tab:laclip_results}).
In this section, we simplify the notation $\mathcal{L}_\mathrm{1} (\mathbf{X}_\mathrm{I}, \mathbf{X}_\mathrm{T})$, $\mathcal{L}_\mathrm{1} (\mathbf{X}_\mathrm{I}, \mathbf{X}_\mathrm{T}^\prime)$, and $\mathcal{L}_\mathrm{1} (\mathbf{X}_\mathrm{I}, \mathbf{X}_\mathrm{T}^{\prime\prime})$ to $\mathcal{L}_\mathrm{1}$, $\mathcal{L}_\mathrm{1}^\prime$, and $\mathcal{L}_\mathrm{1}^{\prime\prime}$, respectively.
Note that our ParaCLIP model was trained using the combined loss functions, $\mathcal{L}_\mathrm{1}^{\prime\prime}$$+$\ltwosim$+$\lthreesim, as detailed in \Cref{subsec:objectives}.

First, we fine-tuned the OpenAI's CLIP model using the same set of image-caption pairs in LAION-400M as our model, excluding paraphrases (referred to as ``\lonesim'').
While there was an overall improvement in performance, it still fell short of our ParaCLIP model's performance.
When $\mathcal{L}_\mathrm{1}^{\prime\prime}$~was omitted (i.g., \ltwosim~$+$~\lthreesim), the model showed the best performance on the VG-R, VG-A, and STS tasks, but the performance on image classification and standard text and image retrieval significantly degraded.
This indicates that $\mathcal{L}_\mathrm{1}^{\prime\prime}$~was crucial in preserving the representations of CLIP acquired during pre-training.
Although simply augmenting training data with synthetic paraphrases (i.e., \lonesim~$+$~$\mathcal{L}_\mathrm{1}^{\prime}$ and \lonesim~$+$~$\mathcal{L}_\mathrm{1}^{\prime} + \mathcal{L}_\mathrm{1}^{\prime\prime}$)~generally led to performance improvements, the improvements in the STS tasks were not substantial compared to the models with the \ltwosim~and~\lthreesim~losses.
Applying~\lthreesim~was particularly effective for STS because it involved comparing pairs of semantically similar ``plain'' text (not pairs of noisy caption and plain text), which aligns well with the goal of STS.
Finally, our ParaCLIP model, incorporating three losses (i.e., $\mathcal{L}_\mathrm{1}^{\prime\prime}$~$+$~\ltwosim~$+$~\lthreesim), showed the most balanced performance across all tasks among the various models evaluated.
In particular, applying $\mathcal{L}_\mathrm{1}^{\prime\prime}$ instead of $\mathcal{L}_\mathrm{1}$ proved to be generally effective.
\section{Conclusion}

In this study, we proposed a two-step paraphrasing approach for enhancing the representations of CLIP for paraphrases that may occur in text inputs in real-world applications.
Our ParaCLIP models, fine-tuned using synthetic paraphrases, outperformed baseline models by a large margin on various tasks requiring language semantics and compositional understanding, including paraphrased retrieval.

\section*{Limitations}
Our method sometimes degrades the performance of CLIP on conventional vision and vision-language tasks such as zero-shot classification and image retrieval.
A significant factor contributing to this performance variation may be the sensitivity of the infoNCE loss to changes in batch size.
We observed consistent improvements in the image classification and text/image retrieval tasks by scaling up the batch size from 256 to 3K.
Unfortunately, due to constraints in computational resources, we were unable to match the batch size to the scale of CLIP hyperparameters (e.g., OpenAI's CLIP was pre-trained using a batch size of 32K). 
As a result, the effect of batch size in causing the observed performance degradation has not been thoroughly validated in this study. 
Although the primary goal of this paper was to showcase the potential improvements in the CLIP model through synthetic paraphrasing and better generalization ability across various input queries, a comprehensive investigation into the factors contributing to performance degradation should be conducted in future research.

\section*{Acknowledgements}
We thank Fabian Caba Heilbron and Donghee Choi for their help and insightful discussions.
This research was supported by (1) National Research Foundation of Korea (NRF-2023R1A2C3004176), (2) ICT Creative Consilience Program through the Institute of Information \& Communications Technology
Planning \& Evaluation(IITP) grant funded by the Korea government(MSIT)(IITP-2024-2020-0-01819), and (3) a Korea University Grant.

\bibliography{anthology,custom}

\begin{thebibliography}{33}
\expandafter\ifx\csname natexlab\endcsname\relax\def\natexlab#1{#1}\fi

\bibitem[{Agirre et~al.(2015)Agirre, Banea, Cardie, Cer, Diab, Gonzalez-Agirre, Guo, Lopez-Gazpio, Maritxalar, Mihalcea, Rigau, Uria, and Wiebe}]{agirre-etal-2015-semeval}
Eneko Agirre, Carmen Banea, Claire Cardie, Daniel Cer, Mona Diab, Aitor Gonzalez-Agirre, Weiwei Guo, I{\~n}igo Lopez-Gazpio, Montse Maritxalar, Rada Mihalcea, German Rigau, Larraitz Uria, and Janyce Wiebe. 2015.
\newblock \href {https://doi.org/10.18653/v1/S15-2045} {{S}em{E}val-2015 task 2: Semantic textual similarity, {E}nglish, {S}panish and pilot on interpretability}.
\newblock In \emph{Proceedings of the 9th International Workshop on Semantic Evaluation ({S}em{E}val 2015)}, pages 252--263, Denver, Colorado. Association for Computational Linguistics.

\bibitem[{Agirre et~al.(2014)Agirre, Banea, Cardie, Cer, Diab, Gonzalez-Agirre, Guo, Mihalcea, Rigau, and Wiebe}]{agirre-etal-2014-semeval}
Eneko Agirre, Carmen Banea, Claire Cardie, Daniel Cer, Mona Diab, Aitor Gonzalez-Agirre, Weiwei Guo, Rada Mihalcea, German Rigau, and Janyce Wiebe. 2014.
\newblock \href {https://doi.org/10.3115/v1/S14-2010} {{S}em{E}val-2014 task 10: Multilingual semantic textual similarity}.
\newblock In \emph{Proceedings of the 8th International Workshop on Semantic Evaluation ({S}em{E}val 2014)}, pages 81--91, Dublin, Ireland. Association for Computational Linguistics.

\bibitem[{Agirre et~al.(2016)Agirre, Banea, Cer, Diab, Gonzalez-Agirre, Mihalcea, Rigau, and Wiebe}]{agirre-etal-2016-semeval}
Eneko Agirre, Carmen Banea, Daniel Cer, Mona Diab, Aitor Gonzalez-Agirre, Rada Mihalcea, German Rigau, and Janyce Wiebe. 2016.
\newblock \href {https://doi.org/10.18653/v1/S16-1081} {{S}em{E}val-2016 task 1: Semantic textual similarity, monolingual and cross-lingual evaluation}.
\newblock In \emph{Proceedings of the 10th International Workshop on Semantic Evaluation ({S}em{E}val-2016)}, pages 497--511, San Diego, California. Association for Computational Linguistics.

\bibitem[{Agirre et~al.(2012)Agirre, Cer, Diab, and Gonzalez-Agirre}]{agirre-etal-2012-semeval}
Eneko Agirre, Daniel Cer, Mona Diab, and Aitor Gonzalez-Agirre. 2012.
\newblock \href {https://aclanthology.org/S12-1051} {{S}em{E}val-2012 task 6: A pilot on semantic textual similarity}.
\newblock In \emph{*{SEM} 2012: The First Joint Conference on Lexical and Computational Semantics {--} Volume 1: Proceedings of the main conference and the shared task, and Volume 2: Proceedings of the Sixth International Workshop on Semantic Evaluation ({S}em{E}val 2012)}, pages 385--393, Montr{\'e}al, Canada. Association for Computational Linguistics.

\bibitem[{Agirre et~al.(2013)Agirre, Cer, Diab, Gonzalez-Agirre, and Guo}]{agirre-etal-2013-sem}
Eneko Agirre, Daniel Cer, Mona Diab, Aitor Gonzalez-Agirre, and Weiwei Guo. 2013.
\newblock \href {https://aclanthology.org/S13-1004} {*{SEM} 2013 shared task: Semantic textual similarity}.
\newblock In \emph{Second Joint Conference on Lexical and Computational Semantics (*{SEM}), Volume 1: Proceedings of the Main Conference and the Shared Task: Semantic Textual Similarity}, pages 32--43, Atlanta, Georgia, USA. Association for Computational Linguistics.

\bibitem[{Brown et~al.(2020)Brown, Mann, Ryder, Subbiah, Kaplan, Dhariwal, Neelakantan, Shyam, Sastry, Askell et~al.}]{brown2020language}
Tom Brown, Benjamin Mann, Nick Ryder, Melanie Subbiah, Jared~D Kaplan, Prafulla Dhariwal, Arvind Neelakantan, Pranav Shyam, Girish Sastry, Amanda Askell, et~al. 2020.
\newblock \href {https://papers.nips.cc/paper/2020/hash/1457c0d6bfcb4967418bfb8ac142f64a-Abstract.html} {Language models are few-shot learners}.
\newblock \emph{Advances in neural information processing systems}, 33:1877--1901.

\bibitem[{Cer et~al.(2017)Cer, Diab, Agirre, Lopez-Gazpio, and Specia}]{cer-etal-2017-semeval}
Daniel Cer, Mona Diab, Eneko Agirre, I{\~n}igo Lopez-Gazpio, and Lucia Specia. 2017.
\newblock \href {https://doi.org/10.18653/v1/S17-2001} {{S}em{E}val-2017 task 1: Semantic textual similarity multilingual and crosslingual focused evaluation}.
\newblock In \emph{Proceedings of the 11th International Workshop on Semantic Evaluation ({S}em{E}val-2017)}, pages 1--14, Vancouver, Canada. Association for Computational Linguistics.

\bibitem[{Cheng et~al.(2024)Cheng, Shin, Vasconcelos, Russell, and Caba~Heilbron}]{cheng2024adapting}
Jiacheng Cheng, Hijung~Valentina Shin, Nuno Vasconcelos, Bryan Russell, and Fabian Caba~Heilbron. 2024.
\newblock Adapting clip to paraphrased retrieval with pretrained language models.

\bibitem[{Cherti et~al.(2023)Cherti, Beaumont, Wightman, Wortsman, Ilharco, Gordon, Schuhmann, Schmidt, and Jitsev}]{cherti2023reproducible}
Mehdi Cherti, Romain Beaumont, Ross Wightman, Mitchell Wortsman, Gabriel Ilharco, Cade Gordon, Christoph Schuhmann, Ludwig Schmidt, and Jenia Jitsev. 2023.
\newblock \href {https://arxiv.org/abs/2212.07143} {Reproducible scaling laws for contrastive language-image learning}.
\newblock In \emph{Proceedings of the IEEE/CVF Conference on Computer Vision and Pattern Recognition}, pages 2818--2829.

\bibitem[{Chuang et~al.(2022)Chuang, Dangovski, Luo, Zhang, Chang, Soljacic, Li, Yih, Kim, and Glass}]{chuang-etal-2022-diffcse}
Yung-Sung Chuang, Rumen Dangovski, Hongyin Luo, Yang Zhang, Shiyu Chang, Marin Soljacic, Shang-Wen Li, Scott Yih, Yoon Kim, and James Glass. 2022.
\newblock \href {https://doi.org/10.18653/v1/2022.naacl-main.311} {{D}iff{CSE}: Difference-based contrastive learning for sentence embeddings}.
\newblock In \emph{Proceedings of the 2022 Conference of the North American Chapter of the Association for Computational Linguistics: Human Language Technologies}, pages 4207--4218, Seattle, United States. Association for Computational Linguistics.

\bibitem[{Conneau et~al.(2017)Conneau, Kiela, Schwenk, Barrault, and Bordes}]{conneau-etal-2017-supervised}
Alexis Conneau, Douwe Kiela, Holger Schwenk, Lo{\"\i}c Barrault, and Antoine Bordes. 2017.
\newblock \href {https://doi.org/10.18653/v1/D17-1070} {Supervised learning of universal sentence representations from natural language inference data}.
\newblock In \emph{Proceedings of the 2017 Conference on Empirical Methods in Natural Language Processing}, pages 670--680, Copenhagen, Denmark. Association for Computational Linguistics.

\bibitem[{Deng et~al.(2009)Deng, Dong, Socher, Li, Li, and Fei-Fei}]{deng2009imagenet}
Jia Deng, Wei Dong, Richard Socher, Li-Jia Li, Kai Li, and Li~Fei-Fei. 2009.
\newblock \href {https://ieeexplore.ieee.org/document/5206848} {Imagenet: A large-scale hierarchical image database}.
\newblock In \emph{2009 IEEE conference on computer vision and pattern recognition}, pages 248--255. Ieee.

\bibitem[{Dosovitskiy et~al.(2021)Dosovitskiy, Beyer, Kolesnikov, Weissenborn, Zhai, Unterthiner, Dehghani, Minderer, Heigold, Gelly, Uszkoreit, and Houlsby}]{dosovitskiy2020image}
Alexey Dosovitskiy, Lucas Beyer, Alexander Kolesnikov, Dirk Weissenborn, Xiaohua Zhai, Thomas Unterthiner, Mostafa Dehghani, Matthias Minderer, Georg Heigold, Sylvain Gelly, Jakob Uszkoreit, and Neil Houlsby. 2021.
\newblock \href {https://openreview.net/forum?id=YicbFdNTTy} {An image is worth 16x16 words: Transformers for image recognition at scale}.
\newblock In \emph{International Conference on Learning Representations}.

\bibitem[{Fagin et~al.(2003)Fagin, Kumar, and Sivakumar}]{fagin2003comparing}
Ronald Fagin, Ravi Kumar, and Dakshinamurthi Sivakumar. 2003.
\newblock \href {https://dl.acm.org/doi/abs/10.5555/644108.644113} {Comparing top k lists}.
\newblock \emph{SIAM Journal on discrete mathematics}, 17(1):134--160.

\bibitem[{Fan et~al.(2023)Fan, Krishnan, Isola, Katabi, and Tian}]{fan2023improving}
Lijie Fan, Dilip Krishnan, Phillip Isola, Dina Katabi, and Yonglong Tian. 2023.
\newblock \href {https://arxiv.org/abs/2305.20088} {Improving clip training with language rewrites}.
\newblock \emph{Advances in Neural Information Processing Systems}.

\bibitem[{Feng et~al.(2023)Feng, He, Fu, Jampani, Akula, Narayana, Basu, Wang, and Wang}]{feng2023training}
Weixi Feng, Xuehai He, Tsu-Jui Fu, Varun Jampani, Arjun Akula, Pradyumna Narayana, Sugato Basu, Xin~Eric Wang, and William~Yang Wang. 2023.
\newblock \href {https://arxiv.org/abs/2212.05032} {Training-free structured diffusion guidance for compositional text-to-image synthesis}.
\newblock \emph{The Eleventh International Conference on Learning Representations}.

\bibitem[{Gao et~al.(2021)Gao, Yao, and Chen}]{gao-etal-2021-simcse}
Tianyu Gao, Xingcheng Yao, and Danqi Chen. 2021.
\newblock \href {https://doi.org/10.18653/v1/2021.emnlp-main.552} {{S}im{CSE}: Simple contrastive learning of sentence embeddings}.
\newblock In \emph{Proceedings of the 2021 Conference on Empirical Methods in Natural Language Processing}, pages 6894--6910, Online and Punta Cana, Dominican Republic. Association for Computational Linguistics.

\bibitem[{Jaccard(1912)}]{jaccard1912distribution}
Paul Jaccard. 1912.
\newblock \href {https://nph.onlinelibrary.wiley.com/doi/10.1111/j.1469-8137.1912.tb05611.x} {The distribution of the flora in the alpine zone. 1}.
\newblock \emph{New phytologist}, 11(2):37--50.

\bibitem[{Lin et~al.(2014)Lin, Maire, Belongie, Hays, Perona, Ramanan, Doll{\'a}r, and Zitnick}]{lin2014microsoft}
Tsung-Yi Lin, Michael Maire, Serge Belongie, James Hays, Pietro Perona, Deva Ramanan, Piotr Doll{\'a}r, and C~Lawrence Zitnick. 2014.
\newblock \href {https://arxiv.org/abs/1405.0312} {Microsoft coco: Common objects in context}.
\newblock In \emph{Computer Vision--ECCV 2014: 13th European Conference, Zurich, Switzerland, September 6-12, 2014, Proceedings, Part V 13}, pages 740--755. Springer.

\bibitem[{Liu et~al.(2019)Liu, Ott, Goyal, Du, Joshi, Chen, Levy, Lewis, Zettlemoyer, and Stoyanov}]{liu2019roberta}
Yinhan Liu, Myle Ott, Naman Goyal, Jingfei Du, Mandar Joshi, Danqi Chen, Omer Levy, Mike Lewis, Luke Zettlemoyer, and Veselin Stoyanov. 2019.
\newblock \href {https://arxiv.org/abs/1907.11692} {Roberta: A robustly optimized bert pretraining approach}.
\newblock \emph{arXiv preprint arXiv:1907.11692}.

\bibitem[{Loshchilov and Hutter(2019)}]{loshchilov2017decoupled}
Ilya Loshchilov and Frank Hutter. 2019.
\newblock \href {https://openreview.net/forum?id=Bkg6RiCqY7} {Decoupled weight decay regularization}.
\newblock In \emph{International Conference on Learning Representations}.

\bibitem[{Marelli et~al.(2014)Marelli, Menini, Baroni, Bentivogli, Bernardi, and Zamparelli}]{marelli-etal-2014-sick}
Marco Marelli, Stefano Menini, Marco Baroni, Luisa Bentivogli, Raffaella Bernardi, and Roberto Zamparelli. 2014.
\newblock \href {http://www.lrec-conf.org/proceedings/lrec2014/pdf/363_Paper.pdf} {A {SICK} cure for the evaluation of compositional distributional semantic models}.
\newblock In \emph{Proceedings of the Ninth International Conference on Language Resources and Evaluation ({LREC}'14)}, pages 216--223, Reykjavik, Iceland. European Language Resources Association (ELRA).

\bibitem[{Oord et~al.(2018)Oord, Li, and Vinyals}]{oord2018representation}
Aaron van~den Oord, Yazhe Li, and Oriol Vinyals. 2018.
\newblock \href {https://arxiv.org/abs/1807.03748} {Representation learning with contrastive predictive coding}.
\newblock \emph{arXiv preprint arXiv:1807.03748}.

\bibitem[{OpenAI(2022)}]{openai2022chatgpt}
OpenAI. 2022.
\newblock \href {https://openai.com/blog/chatgpt} {Introducing chatgpt}.

\bibitem[{Plummer et~al.(2015)Plummer, Wang, Cervantes, Caicedo, Hockenmaier, and Lazebnik}]{plummer2015flickr30k}
Bryan~A Plummer, Liwei Wang, Chris~M Cervantes, Juan~C Caicedo, Julia Hockenmaier, and Svetlana Lazebnik. 2015.
\newblock \href {https://arxiv.org/abs/1505.04870} {Flickr30k entities: Collecting region-to-phrase correspondences for richer image-to-sentence models}.
\newblock In \emph{Proceedings of the IEEE international conference on computer vision}, pages 2641--2649.

\bibitem[{Radford et~al.(2021)Radford, Kim, Hallacy, Ramesh, Goh, Agarwal, Sastry, Askell, Mishkin, Clark et~al.}]{radford2021learning}
Alec Radford, Jong~Wook Kim, Chris Hallacy, Aditya Ramesh, Gabriel Goh, Sandhini Agarwal, Girish Sastry, Amanda Askell, Pamela Mishkin, Jack Clark, et~al. 2021.
\newblock \href {https://arxiv.org/abs/2103.00020} {Learning transferable visual models from natural language supervision}.
\newblock In \emph{International conference on machine learning}, pages 8748--8763. PMLR.

\bibitem[{Reimers and Gurevych(2019)}]{reimers-gurevych-2019-sentence}
Nils Reimers and Iryna Gurevych. 2019.
\newblock \href {https://doi.org/10.18653/v1/D19-1410} {Sentence-{BERT}: Sentence embeddings using {S}iamese {BERT}-networks}.
\newblock In \emph{Proceedings of the 2019 Conference on Empirical Methods in Natural Language Processing and the 9th International Joint Conference on Natural Language Processing (EMNLP-IJCNLP)}, pages 3982--3992, Hong Kong, China. Association for Computational Linguistics.

\bibitem[{Rombach et~al.(2022)Rombach, Blattmann, Lorenz, Esser, and Ommer}]{rombach2022high}
Robin Rombach, Andreas Blattmann, Dominik Lorenz, Patrick Esser, and Bj{\"o}rn Ommer. 2022.
\newblock \href {https://arxiv.org/abs/2112.10752} {High-resolution image synthesis with latent diffusion models}.
\newblock In \emph{Proceedings of the IEEE/CVF conference on computer vision and pattern recognition}, pages 10684--10695.

\bibitem[{Saharia et~al.(2022)Saharia, Chan, Saxena, Li, Whang, Denton, Ghasemipour, Gontijo~Lopes, Karagol~Ayan, Salimans et~al.}]{saharia2022photorealistic}
Chitwan Saharia, William Chan, Saurabh Saxena, Lala Li, Jay Whang, Emily~L Denton, Kamyar Ghasemipour, Raphael Gontijo~Lopes, Burcu Karagol~Ayan, Tim Salimans, et~al. 2022.
\newblock \href {https://arxiv.org/abs/2205.11487} {Photorealistic text-to-image diffusion models with deep language understanding}.
\newblock \emph{Advances in Neural Information Processing Systems}, 35:36479--36494.

\bibitem[{Schuhmann et~al.(2022)Schuhmann, Beaumont, Vencu, Gordon, Wightman, Cherti, Coombes, Katta, Mullis, Wortsman et~al.}]{schuhmann2022laion}
Christoph Schuhmann, Romain Beaumont, Richard Vencu, Cade Gordon, Ross Wightman, Mehdi Cherti, Theo Coombes, Aarush Katta, Clayton Mullis, Mitchell Wortsman, et~al. 2022.
\newblock \href {https://arxiv.org/abs/2210.08402} {Laion-5b: An open large-scale dataset for training next generation image-text models}.
\newblock \emph{Advances in Neural Information Processing Systems}, 35:25278--25294.

\bibitem[{Schuhmann et~al.(2021)Schuhmann, Vencu, Beaumont, Kaczmarczyk, Mullis, Katta, Coombes, Jitsev, and Komatsuzaki}]{schuhmann2021laion}
Christoph Schuhmann, Richard Vencu, Romain Beaumont, Robert Kaczmarczyk, Clayton Mullis, Aarush Katta, Theo Coombes, Jenia Jitsev, and Aran Komatsuzaki. 2021.
\newblock \href {https://arxiv.org/abs/2111.02114} {Laion-400m: Open dataset of clip-filtered 400 million image-text pairs}.
\newblock \emph{NeurIPS Data-Centric AI Workshop 2021}.

\bibitem[{Touvron et~al.(2023)Touvron, Lavril, Izacard, Martinet, Lachaux, Lacroix, Rozi{\`e}re, Goyal, Hambro, Azhar et~al.}]{touvron2023llama}
Hugo Touvron, Thibaut Lavril, Gautier Izacard, Xavier Martinet, Marie-Anne Lachaux, Timoth{\'e}e Lacroix, Baptiste Rozi{\`e}re, Naman Goyal, Eric Hambro, Faisal Azhar, et~al. 2023.
\newblock \href {https://arxiv.org/abs/2302.13971} {Llama: Open and efficient foundation language models}.
\newblock \emph{arXiv preprint arXiv:2302.13971}.

\bibitem[{Yuksekgonul et~al.(2023)Yuksekgonul, Bianchi, Kalluri, Jurafsky, and Zou}]{yuksekgonul2022and}
Mert Yuksekgonul, Federico Bianchi, Pratyusha Kalluri, Dan Jurafsky, and James Zou. 2023.
\newblock \href {https://arxiv.org/abs/2210.01936} {When and why vision-language models behave like bags-of-words, and what to do about it?}
\newblock In \emph{The Eleventh International Conference on Learning Representations}.

\end{thebibliography}
\bibliographystyle{acl_natbib}

\clearpage

\appendix

\section{Implementation Details}
\label{appendix:details}

In the data generation process, we used the \texttt{gpt-35-turbo-0301} model with the temperature of 1.0 and top-p of 0.1.
We paid approximately 130 USD for using ChatGPT to generate 300K paraphrases for captions and 300K additional paraphrases for generated paraphrases.

We used the checkpoints of CLIP models provided in the official OpenCLIP GitHub repository.\footnote{{https://github.com/mlfoundations/open\_clip}}
We used \texttt{openai} for OpenAI's CLIP, \texttt{laion400m\_e32} for OpenCLIP (400M), \texttt{laion2b\_s34b\_b79k} for OpenCLIP (2B), and \texttt{laion2b\_s12b\_b32k} for OpenCLIP-RoBERTa.
Our ParaCLIP models were trained for one epoch using the AdamW optimizer~\cite{loshchilov2017decoupled}, coupled with a cosine annealing scheduler, on eight A100 80G GPUs. 
For fine-tuning, a learning rate of 5e-7, a batch size of 3,072, and a weight decay rate of 0.001 were used. 
All reported scores were measured on a single run.

\section{Metrics in Paraphrased Retrieval}
\label{appendix:metrics}

\paragraph{Average overlap}
The top-k average overlap (AO@k)~\cite{fagin2003comparing} quantifies the rank similarity between the top-k elements of the two lists. 
Let $L_\mathrm{a}$ and $L_\mathrm{b}$ be ordered lists of retrieved images for two different queries. 
AO@k between the two lists is calculated based on the weighted sum of intersections of truncated lists as follows: 
\begin{equation}
    \mathrm{AO@k}(L_\mathrm{a}, L_\mathrm{b}) := \frac{1}{k}\sum_{d=1}^k \frac{|L_\mathrm{a}^d \cap L_\mathrm{b}^d|}{d},
\end{equation}
where $L^d_\mathrm{a}= L_\mathrm{a}[1 : d]$ and $L^d_\mathrm{b}
= L_\mathrm{b}[1 : d]$ represent the truncated lists at depth $d$ and $|L^d_\mathrm{a} \cap L^d_\mathrm{b}|$ indicates the cardinality of the set intersection between these truncated lists.
When AO@k equals 1, it means that the top-k elements of $L_\mathrm{a}$ and $L_\mathrm{b}$ are exactly the same. 
Conversely, when AO@k equals 0, it implies that there is no overlap whatsoever between the top-k elements of $L_\mathrm{a}$ and $L_\mathrm{b}$.
AO@k gives more weight to the higher-ranked retrieval results because they contribute to more terms in the overall summation compared to lower-ranked results.

\paragraph{Jaccard similarity}
The top-k Jaccard similarity (JS@k)~\cite{jaccard1912distribution} is calculated as the ratio of the intersection to the union of the top-k elements in two lists as follows:
\begin{equation}
    \mathrm{JS@k}(L_\mathrm{a}, L_\mathrm{b}) := \frac{|L_\mathrm{a}^k \cap L_\mathrm{b}^k|}{|L_\mathrm{a}^k \cup L_\mathrm{b}^k|},
\end{equation}
where $|L^k_\mathrm{a} \cup L^k_\mathrm{b}|$ is the cardinality of the set union between $L^k_\mathrm{a}$ and $L^k_\mathrm{b}$.
JS@k equals 0 when $L^k_\mathrm{a}$ and $L^k_\mathrm{b}$ are disjoint and equals 1 when $L^k_\mathrm{a}$ and $L^k_\mathrm{b}$ contain the same retrieval results (although not necessarily in the same order). 
Unlike the average overlap, the Jaccard similarity does not assign more weight to the higher-ranked retrieval results. 


\section{Case Study}
\label{appendix:case_study}

\begin{figure*}[t]
\centering
\begin{subfigure}{\linewidth}
  \centering
  \includegraphics[width=\linewidth]{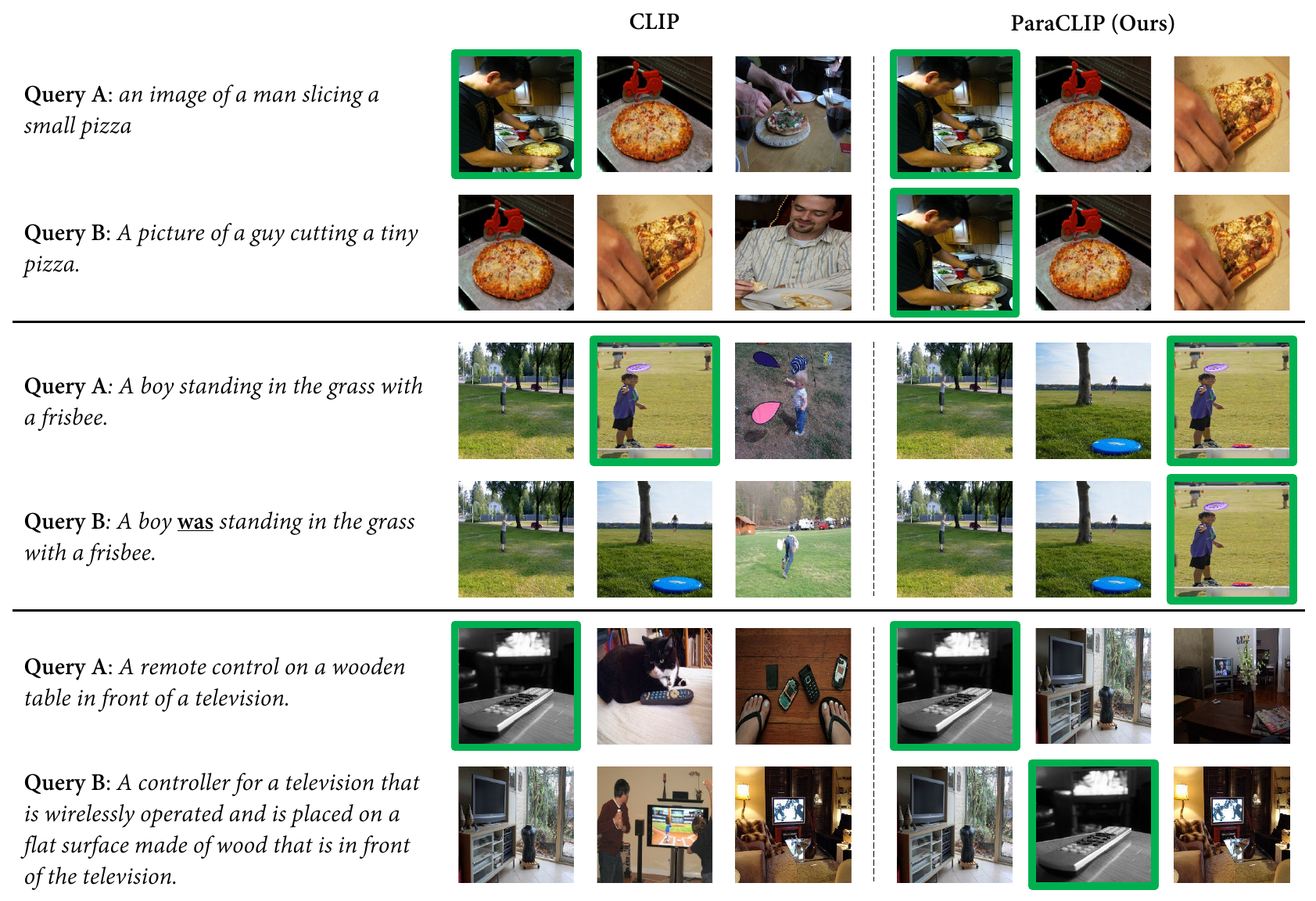} 
\end{subfigure}

\caption{
Examples of retrieved images by the CLIP~\cite{radford2021learning} and our ParaCLIP models for two different queries. 
Note that the queries are obtained from the paraphrased retrieval dataset, and query B is a paraphrase for query A.
The gold images are denoted by a bold border.
}
\label{fig:case_study}
\end{figure*}

\Cref{fig:case_study}~shows several examples where our ParaCLIP model yieled better retrieval results than OpenAI's CLIP for paraphrased queries.
In the first example, the paraphrased query (query B) contained several synonyms such as ``picture,'' ``guy,'' ``cutting,'' and ``tiny,'' replacing the words ``image,'' ``man,'' ``slicing,'' and ``small,'' respectively.
While the CLIP model output dissimilar results for the given two queries, resulting in a performance drop for query B, ParaCLIP consistently produced identical results for both queries.
In the second example, the only difference between the queries was the word ``was.''
Despite this minor variation, CLIP generated different sets of images.
On the other hand, ParaCLIP returned the same images for both queries and achieved a better recall for query B, although the recall score for query A was slightly lower than that of CLIP.
In the last example, query B was created by expanding the short query A into longer expressions.
For instance, the concise phrase ``a remote control'' was transformed into the more elaborate phrase ``a controller for a television that is wirelessly operated.''
While CLIP exhibited high sensitivity to this long paraphrased query, ParaCLIP demonstrated greater robustness, resulting in more consistent results and superior recall scores.

\end{document}